# FishAI 2.0: Marine Fish Image Classification with Multi-modal Few-shot Learning

Chenghan Yang[1#], Peng Zhou[1,2,3*], Dong-Sheng Zhang[2,3], Yueyun Wang[2,3], Hong-Bin Shen[1], Xiaoyong Pan[1*]

[1] Institute of Image Processing and Pattern Recognition, Shanghai Jiao Tong University, and Key Laboratory of System Control and Information Processing, Ministry of Education of China.

[2] State Key Laboratory of Submarine Geoscience, Second Institute of Oceanography, Ministry of Natural Resources, Hangzhou, China

[3] Key Laboratory of Marine Ecosystem Dynamics, Ministry of Natural Resources and Second Institute of Oceanography, Ministry of Natural Resources, Hangzhou, China

*Corresponding authors: X. Pan, 2008xypan@sjtu.edu.cn or P. Zhou, zhoupeng@sio.org.cn

## Abstract

Traditional marine biological image recognition faces challenges of incomplete datasets and unsatisfied model accuracy, particularly for few-shot conditions of rare species where data scarcity significantly hampers the performance. To address these issues, this study proposes an intelligent marine fish recognition framework, FishAI 2.0, integrating multimodal few-shot deep learning techniques with image generation for data augmentation. First, a hierarchical marine fish benchmark dataset, which provide a comprehensive data foundation for subsequent model training, is utilized to train FishAI 2.0 model. To address the data scarcity of rare classes, the large language model DeepSeek was employed to generate high-quality textual descriptions, which are input into Stable Diffusion 2 for image augmentation through a hierarchical diffusion strategy that extracts latent encoding to construct a multimodal feature space. The enhanced visual–textual datasets were then fed into a Contrastive Language-Image Pre-Training (CLIP) based model, enabling robust few-shot image recognition. Experimental results demonstrate that FishAI 2.0 achieves a Top-1 accuracy of 91.67% and Top-5 accuracy of 97.97% at the family level, outperforming baseline CLIP and ViT models with a large margin for the minority classes with the number of training samples fewer than 10. To better apply FishAI 2.0 to real world, at the genus and species level, FishAI 2.0 respectively achieves a Top-1 accuracy of 87.58% and 85.42%, demonstrating the practical utility. In summary, FishAI 2.0 improves the efficiency and accuracy of marine fish identification and provides a scalable technical solution for marine ecological monitoring and conservation, highlighting its scientific value and practical applicability.

## 1 Introduction

Ecologically, fish (Pisces) are key components of aquatic ecosystems, mediating the energy transfer and nutrient cycling through food webs. In modern fish ecology research, the dual threats of invasive and endangered species have gained increasing attention. Xu et al. conducted a global-scale analysis showing that freshwater fish invasions are more likely when non-native species are phylogenetically close to native ones, suggesting that ecological compatibility facilitates the establishment[1]. The silver carp (*Hypophthalmichthys molitrix*), one of North America's most pervasive invasive species, demonstrates physiological adaptations at invasion fronts. Jeffrey et al. documented enhanced detoxification activity and altered metabolic processes, reflecting environmental stressors that may constrain upstream spread[2]. Meanwhile, endangered taxa such as the red-finned blue-eye (*Scaturiginichthys vermeilipinnis*) from central Queensland require precise monitoring to prevent extinction[3]. However, detecting invasive and endangered species by human is costly and time-consuming, requiring strong expert knowledge in marine biology.

To solve the problem, one important way is to build deep learning models for automatic fish recognition, which generally requires a large training image dataset. Recently, fish dataset limitations remain a major bottleneck, public datasets such as Fish4Knowledge (23 species)[4] and ImageCLEF (12 species)[5] have restricted taxonomic coverage. Some researchers have built proprietary datasets, such as market-sourced collections, but annotation inconsistencies persist. Especially, rare species are often underrepresented, leading to sharp performance drop for rare fishes. Moreover, with relatively comprehensive datasets, the class imbalance remains problematic, particularly for challenging imaging conditions. For example, Yang et al. reported that in the WoRMS dataset (300 species), reducing per-class training samples from over 20 to fewer than 20 caused the accuracy of FishAI to decrease from 92.5% to 68.3%[6]. Thus, a more data-balanced and comprehensive dataset needs be constructed. In 2023, Khan et. al generate FishNet dataset, which is a large-scale, structured marine biological image dataset designed to support tasks such as fish recognition, detection, and functional trait prediction, with particular suitability for few-shot learning and ecological research[7]. This dataset offers several distinctive advantages. 1) taxonomic completeness: FishNet strictly follows the taxonomic standards of the global fish database, covering the full hierarchy of *Family*, *Genus*, *Species*. 2) controllable image quality: all images were reviewed by professional biologists to ensure that key morphological traits are clearly identifiable. 3) rich metadata: each image is accompanied by 23 ecological attributes, including collection time, geographic location, and water depth. Compared with the limited datasets commonly used in previous studies, such as Fish4Knowledge and ImageCLEF, FishNet demonstrates the advantages in both species' coverage and data quality.

Enlarging datasets does solve the problems for some uncommon species and threatened fishes, but it is difficult to collect a large number of images for extremely rare fishes. When species are rare or newly observed, large annotated datasets are often unavailable. While prior systems, like FishAI, advanced hierarchical classification across taxonomic levels using Vision Transformers, their dependence on large datasets limited the performance at fine-grained levels (genus, species). This highlights the requirement for advanced systems that integrate multimodal information to overcome the data scarcity, improving cross-habitat generalization, and delivering reliable image recognition under few-shot conditions.

Few-shot learning (FSL) offers a promising pathway to address such challenges by enabling robust classification with minimal labeled data, a scenario common in marine research due to difficulties in acquiring and annotating rare species images. In long-tailed datasets, rare taxa have few samples, however conventional deep learning methods, which depend on large-scale annotations, perform poorly and generalize inadequately[8]. FSL techniques, inspired by human learning, employ strategies, such as model transfer, feature reconstruction and semantic expansion, to facilitate learning new categories from only a few examples. Various approaches have been explored to enhance FSL performance, including data augmentation, meta-learning, and multimodal learning. Villon et al. provided a seminal benchmark comparing DL and FSL methods, demonstrating that FSL significantly outperforms conventional DL when only a few samples are available, thereby highlighting annotation bottlenecks in field ecology[9]. Building on this, metric-learning and attention-enhanced architectures have improved fine-grained discrimination under few-shot setting. For instance, the Sandwich Attention Covariance Metric Network (SACovaMNet) augmented CovaMNet with global-local attention to expand inter-class prototype distances, consistently surpassing prior methods on marine datasets[10]. Similarly, improved prototypical networks refined class prototypes and episode design, achieving notable gains across Fish4Knowledge and Croatian Fish datasets[11]. Beyond metric refinement, complementary strategies address the label scarcity and domain adaptation. Few-shot domain-adaptive detectors allow transfer from source to novel target habitats with minimal labels, while attention-based active learning reduces annotation costs without sacrificing the accuracy on fisheries imagery[12, 13]. Transfer-learning baselines such as transferable CNNs and Transformer variants like Fish-TViT remain competitive when moderate data are available, yet their accuracy drops in few-shot setting[14]. Collectively, these studies underscore the potential of few-shot recognition, but also reveal persistent challenges: the reliance on single-modal visual cues, restricted dataset scope (often less than 25 species), sensitivity to underwater conditions (turbidity, lighting), and limited generalization across habitats and capture methods.

While traditional data augmentation and neural network is limited to dataset-level improvements, multimodal pre-trained models such as CLIP offer promising advantages. CLIP Trained via contrastive learning on large-scale image-text pairs acquire rich joint visual-linguistic representations, enabling zero-shot or few-shot classification through textual prompts without additional training. The robust generalization makes CLIP well-suited for marine fish recognition under few-shot constraints, often outperforming state-of-the-art meta-learning methods[15]. Furthermore, coupling CLIP with generative models such as Stable Diffusion can produce high-quality synthetic datasets, improving the classification accuracy[16]. For example, Zhou et al. demonstrated that their FLIER framework, combining text-guided image generation with contrastive learning, improved 1-shot recognition accuracy by 18.7%[17]. Furthermore, Zhou et al. used the similar method to apply it to Decapoda species, demonstrating average accuracies of 0.717 for family, 0.719 for genus, and 0.773 for species[18].

Despite these promising developments, the systematic application to marine fish recognition remains limited. This study aims to integrate CLIP-based multimodal learning with targeted data augmentation to develop a high-efficiency, high-accuracy few-shot fish recognition framework, tailored to the unique characteristics and challenges of marine biodiversity datasets.

In summary, this study makes the following contributions:

1) **Integration of multimodal pre-training with data augmentation for marine fish recognition:** We propose a novel framework named FishAI 2.0, combining CLIP-based multimodal representation learning with generative Stable Diffusion–based data augmentation, enabling the robust recognition under few-shot conditions.

2) **Domain-specific adaptation of CLIP for rare and long-tailed fish species:** We adapt and evaluate FishAI 2.0's few-shot capabilities on marine biodiversity datasets, addressing the challenges of taxonomic imbalance and limited samples of rare species.

3) **Comprehensive benchmark and empirical validation, especially under few-shot conditions:** We construct and benchmark FishAI 2.0, demonstrating the superior accuracy and efficiency compared to state-of-the-art meta-learning and CNN-based baselines, while reducing the requirement for large-scale annotations.

# 2 Material and Method

## 2.1 Overview of FishAI 2.0 model

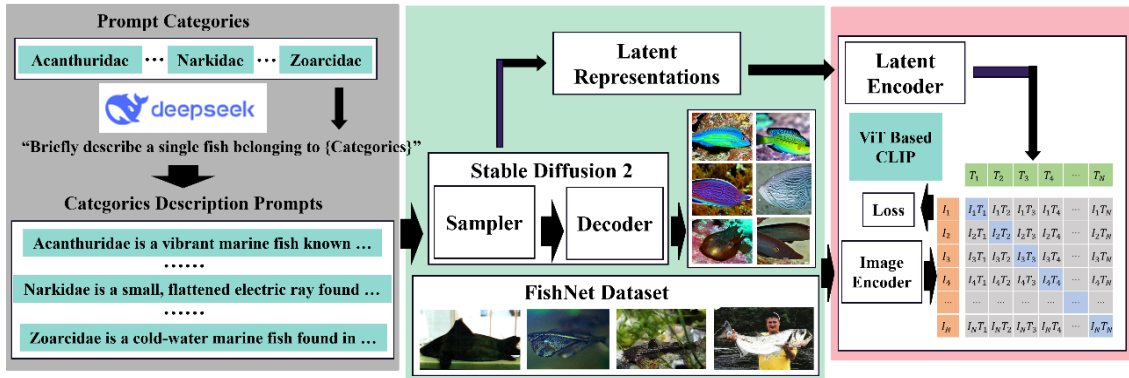

Fig 1. The architecture of the FishAI 2.0. It mainly consists of 3 modules. Firstly, we need to generate prompts for stable diffusion 2 dataset generation via DeepSeek. Then by using the prompts, we use stable diffusion 2 to obtain generated images and latent representations, which are used as the image encoder and latent encoder. Finally, the CLIP-L14 model calculates the training loss, aligning multimodal features from real and synthetic data. This integrated workflow improves the robustness in few-shot fine-grained fish classification.

As shown in Figure 1, this study focuses on integrating Stable Diffusion 2 with the CLIP model to enhance multimodal understanding and generative capabilities for fish recognition. To achieve this, the first step was preparing high-quality textual data. The DeepSeek model was employed to generate prompts in the format *"Briefly describe a single fish belonging to {Categories}"*, from which a large number of semantically rich and structurally text descriptions were extracted. These descriptions not only exhibit strong linguistic expressiveness but also possess high potential for guiding image generation[19]. Subsequently, the collected high-quality texts were input into Stable Diffusion 2 to perform text-to-image generation, producing a substantial number of synthetic fish images that effectively enlarged and balanced the dataset. During the generation process, Stable Diffusion 2 also provided intermediate latent encodings, which capture the intrinsic correspondence between textual and visual representations. These latent codes were then incorporated into the text encoder as additional inputs, serving as a foundation for subsequent CLIP training. Finally, the synthetic images and latent features obtained were fused with the images and labels from the FishNet dataset to construct a multimodal training set that is both diverse and accurate. The CLIP model was trained on this enriched dataset, enabling robust few-shot image recognition and classification.

## 2.2 Data collection

In this study, we choose FishNet as our dataset to train FishAI 2.0. FishNet fully accounts for ecological research requirements, providing a comprehensive benchmark dataset that encourages the development of more accurate and effective tools for monitoring and conserving aquatic ecosystems. For experimental setup, images from each taxonomic level and category were split into a training set (80%) and a test set (20%). This division ensures the model generalization to both known and unseen environments. Specifically, the training set contains 73,145 images, while the test set contains 12,891 images, with no overlap in camera sources between the two sets to ensure reliable evaluation across different environments. The data processing included resizing all images to a fixed dimension, applying normalization to the model input requirements, and encoding category labels into numerical indices.

FishNet annotates images at the family, genus and species level. For the purposes of this study, we primarily focused on the family levels, since it offers high data reliability while reflecting characteristics of long-tailed distributions and few-shot scenarios. With 570 categories at this level, FishNet provides imbalanced yet biologically meaningful data distributions, making it a robust foundation for scientific research. As shown in Table 1, FishAI 2.0 contains more rich categories and images than FishAI 1.0 based on WoRMS from the family level to species level.

Table 1 Data statistics after cleaning the images

| Dataset | Hierarchy | Family | Genus | Species |
|---|---|---|---|---|
| FishAI 2.0 | # of categories | 570 | 3951 | 17357 |
| (FishNet) | # of images | 94532 | 94532 | 94532 |
| FishAI 1.0 | # of categories | 154 | 438 | 808 |
| (WoRMS) | # of images | 2585 | 2558 | 2474 |

## 2.3 DeepSeek-Based Text Generator

In this study, we introduce the DeepSeek model for generating textual prompts for fish image synthesis, aiming to improve the quality of text guidance on few-shot image generation tasks. DeepSeek is an open-source large language model, it adopts a Mixture-of-Experts (MoE) framework, which integrates multi-head latent attention and multi-token prediction mechanisms, and incorporates reinforcement learning strategies. In this project, we mainly leverage the LLM to generate fine-grained textual descriptions of various fish species. Its advantages lie not only in better domain-specific adaptability but also in lower usage costs compared with the GPT-4 API[20]. Furthermore, DeepSeek-R1's reinforcement learning integration and pretraining on diverse corpora make it suitable for handling a wide range of tasks.

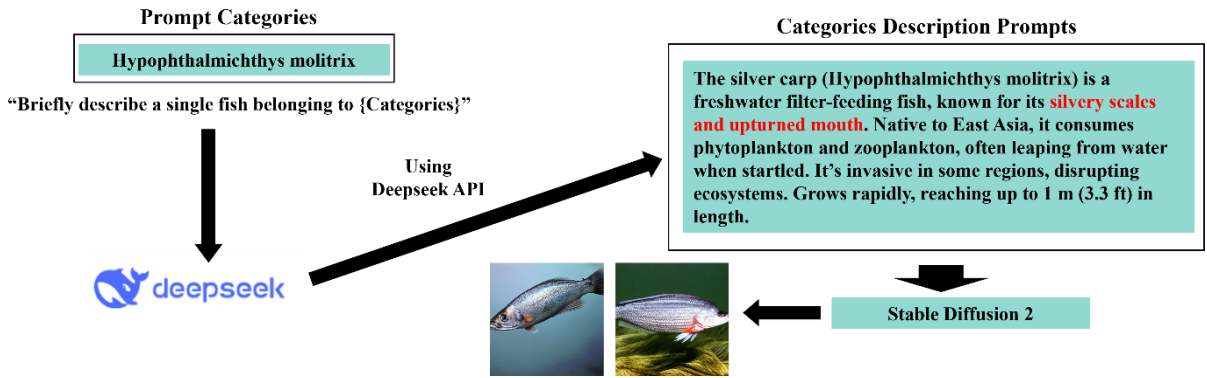

Fig 2. DeepSeek-based Text Label Generator. For example, we ask Deepseek that *"Briefly describe a single fish belonging to Hypophthalmichthys molitrix."* (*Hypophthalmichthys molitrix* can be replaced by other species name for other specie's image generation). Then Deepseek will generate a response, which can be used to image generation for Stable Diffusion 2.

As shown in Figure 2, to generate high-quality fish image descriptions, we designed a specific prompt template: *"Briefly describe a single fish belonging to {Categories}."* Here, *{Categories}* includes different taxonomic levels such as order, family, genus, and species. For example, when inputting the family name

*Hypophthalmichthys molitrix*, the DeepSeek produces the following descriptive prompt: *"The silver carp (Hypophthalmichthys molitrix) is a freshwater filter-feeding fish, known for its silvery scales and upturned mouth. Native to East Asia, it consumes phytoplankton and zooplankton, often leaping from water when startled. It's invasive in some regions, disrupting ecosystems. Grows rapidly, reaching up to 1 m (3.3 ft) in length."* Such descriptive prompts are comprehensive, rich in visual details, and highly effective in guiding image generation models. Therefore, we adopted DeepSeek as the final choice for our text generator. After obtaining the textual descriptions, we feed them into an image generation model, producing high-quality fish images. This integration of natural language processing and image generation not only enhances image quality under few-shot conditions, but also provides novel technical approaches for fish recognition and ecological research.

### 2.4 Stable Diffusion 2-Based Fish Image Generation

As shown in Figure 5, Stable Diffusion 2 (SD2) is a diffusion-based text-to-image generation framework. Its core principle is to transform Gaussian noise into high-quality images through a gradual denoising reverse process. SD2 adopts a hierarchical diffusion structure, combined with a cross-attention mechanism to achieve semantic alignment between text and images. In this study, the input to SD2 is a standardized fish description generated by the DeepSeek model, and the output is a high-fidelity synthetic image along with its corresponding latent space encoding. The generation process proceeds as follows: first, the built-in CLIP text encoder in SD2 maps the input text (e.g., *"Briefly describe a single fish belonging to {Categories}"*) into a 768-dimensional semantic vector, which serves as the conditional guidance signal for the diffusion process. Next, a multi-step denoising operation is carried out in the latent space, where a U-Net architecture predicts noise at each step. Each layer integrates text and image features through cross-attention modules, ensuring that the generated content aligns with the textual description. Finally, the denoised latent vector is reconstructed into an RGB image by the decoder of a variational autoencoder, with a resolution of 512×512. This resolution balances detail preservation with computational efficiency[21].

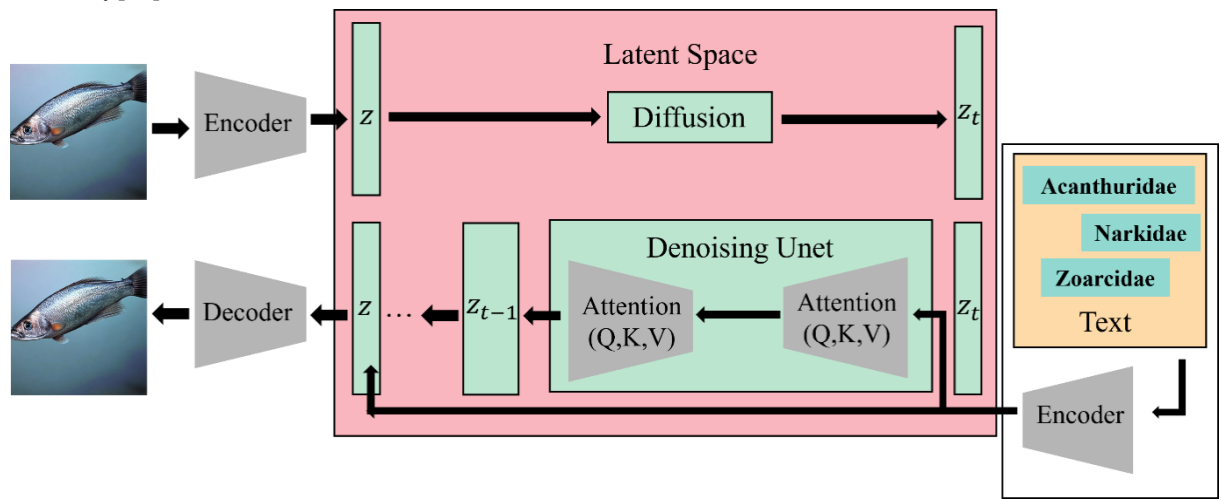

Fig. 3: Overview of the Stable Diffusion 2 Model. Input text is processed by the CLIP text encoder, producing a text embedding. Then random latent vector (noise) is sampled in the compressed latent space. After that, U-Net iteratively removes noise step by step, guided by the text embedding and classifier-free guidance to align with the prompt. Finally, the denoised latent is decoded by the VAE decoder into pixel space, generating the final image.

## 2.5 Few-shot Image Recognition Model FishAI 2.0 Based on CLIP

The CLIP (Contrastive Language–Image Pretraining) model adopts a dual-encoder architecture consisting of an image encoder and a text encoder, achieving cross-modal feature alignment through contrastive learning. By pretraining on large-scale image–text pairs, CLIP learns joint cross-modal semantic representations. Image features are extracted by the image encoder, while text features are extracted by the text encoder. These two encoders are jointly trained using a contrastive loss, bringing semantically related images and texts closer in the vector space, while pushing apart unrelated pairs. In specific tasks such as few-shot fish image recognition, CLIP can be adapted via fine-tuning. Typically, the early layers of the encoders are frozen, while task-specific parameters are updated with a small learning rate and task-relevant loss on the target dataset.

In this study, we fine-tune a pretrained CLIP model to adapt it to the few-shot marine fish classification task. The image encoder employs a Vision Transformer (ViT-B/16) architecture, with input images resized to 224×224 . The text encoder uses a Transformer architecture to process text descriptions generated by DeepSeek. And feature dimensions are unified to 512 for subsequent contrastive similarity calculation using the InfoNCE loss.

To combine the Stable Diffusion 2 and CLIP together, as shown in Fig. 4, SD2 is first used to generate augmented fish images from limited training datasets, producing latent representations that enrich the feature space and compensate for data scarcity. These latent representations, together with real fish images, are then encoded by the CLIP-L14 model, which aligns visual features with corresponding textual prompts across multiple classes. The joint encoding process ensures that both image and latent features are projected into a shared semantic space, enabling robust cross-modal matching. The final loss function is computed over all image–text pairs, driving the model to maximize the similarity between correct matches while minimizing irrelevant associations. This combination of SD2 and CLIP thus provides a powerful strategy: SD2 mitigates few-shot limitations through high-quality synthetic augmentation, while CLIP ensures semantic alignment between visual and textual modalities, resulting in the improved recognition accuracy for rare or fine-grained fish species. After multiple rounds of training and evaluation, the selected CLIP model parameters are listed in Table 2.

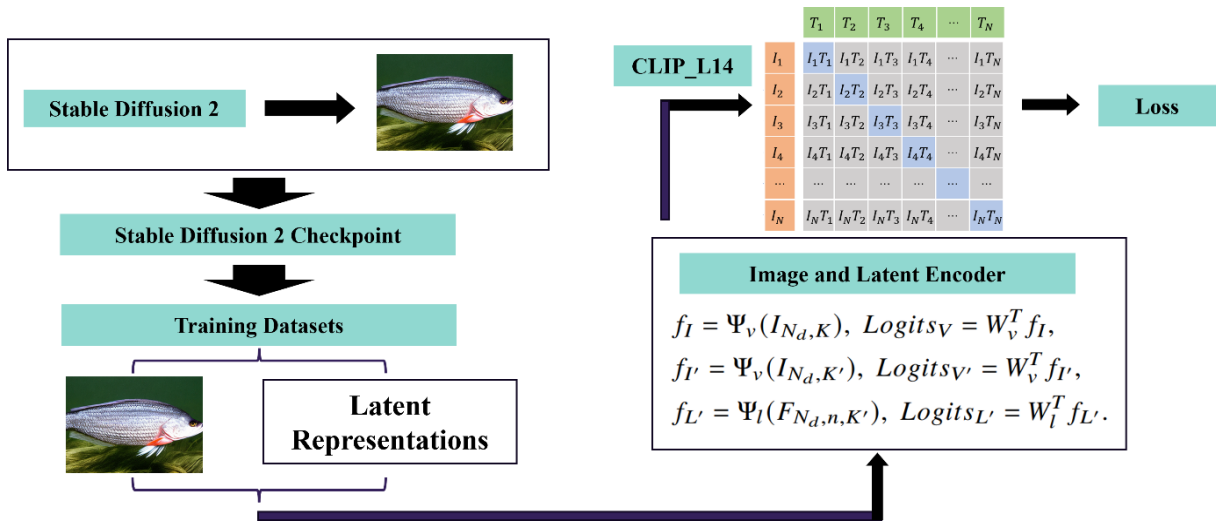

Fig. 4: Architecture of the SD2-CLIP Framework for Few-Shot Marine Fish Recognition. By using the checkpoint of Stable Diffusion 2, we combined the training datasets generated by Stable Diffusion 2 with latent representations fed into to the image and latent encoder.

Table 2. CLIP Model Parameters in FishAI 2.0

| Feature | Parameter |
|---|---|
| Fine-tune | True |
| Batch size | 16 |
| Learning rate | 1e-4 |
| Warmup epochs | 5 |
| Epochs | 100 |
| Learning Decay | 0.7 |
| Weight Decay | 0.05 |
| Base Model | ViT-L/14 |

## 2.6 Evaluation metrics

We adopted a cross-validation strategy by splitting the dataset into the training and test sets with a ratio of 8:2[22]. For classes with fewer than 10 images, samples were divided proportionally at random. The model performance was evaluated using two key metrics: accuracy and cross-entropy loss, which measure classification correctness and prediction reliability, respectively. Additionally, to account for the prediction uncertainty, we reported top-k accuracy (k=2,3,4,5), which evaluates whether the true label appears within the top-k predicted categories.

# 3 Results

## 3.1 Prompt Generation Model Based on DeepSeek

In the field of artificial intelligence, text-to-image generation has achieved remarkable progress in recent years. As an advanced generative model, Stable Diffusion 2 can produce high-quality images from natural language descriptions. However, the quality and accuracy of the generated results are highly dependent on the input prompts. While previous studies have predominantly adopted ChatGPT as the foundation for prompt generation, DeepSeek, as an emerging large language model, demonstrates several advantages over ChatGPT in producing prompts suitable for text-to-image models such as Stable Diffusion 2. In this study, we designed an automated method for generating prompts for Stable Diffusion 2 based on the DeepSeek model. By utilizing fish description data, we evaluated the effectiveness of the generated results.

Table 3. Prompt Generation Cases by DeepSeek

| Family | Prompt Generation |
|---|---|
| *Pristiophoridae* | The sawshark, a member of the Pristiophoridae family, is a unique fish characterized by its elongated, saw-like rostrum lined with sharp teeth. Found in temperate and tropical oceans, it uses its rostrum to detect and slash prey. Typically reaching 1-1.5 meters in length, sawsharks have slender bodies, two dorsal fins, and barbels near their mouths. They are bottom-dwellers, feeding on small fish and invertebrates, and are distinct from sawfish due to their gill slits on the sides of their heads. |
| *Hemiscylliidae* | The epaulette shark, a member of Hemiscylliidae, is a small, slender shark reaching about 1 meter in length. It inhabits shallow coral reefs and tidal pools in the Indo-Pacific. Known for its distinctive black ocellus (eye-like spot) above its pectoral fin, it uses its muscular pectoral fins to \"walk\" across the seafloor. This nocturnal feeder preys on small invertebrates and fish, showcasing remarkable adaptability to low-oxygen environments during tidal changes. |
| *Chilodontidae* | The Chilodontidae family includes small, colorful freshwater fish like the headstanders. One notable species is Chilodus punctatus, the spotted headstander. It has a silvery body with dark spots, a slightly arched back, and a distinctive head-down swimming posture. Native to South American rivers, it thrives in slow-moving, vegetated waters. Omnivorous, it feeds on algae, small invertebrates, and plant matter. Peaceful and social, it's popular in aquariums, requiring soft, acidic water and a well-planted environment. |

The images generated by Stable Diffusion 2 closely matched the original descriptions, validating the effectiveness of the proposed prompt generation approach. For instance, the sawshark image accurately represented the saw-like rostrum and benthic habitat, the butterflyfish image highlighted vivid coloration and coral reef details, and the deep-sea anglerfish image successfully captured bioluminescent features in dark environments. Moreover, comparative experiments revealed that structured prompts generated by DeepSeek yielded images with richer details and stronger contextual coherence than those derived from simple prompts. This automated procedure, leveraging DeepSeek's semantic understanding capabilities, produces prompts that are more faithful to natural language descriptions, thereby enhancing the expressiveness and realism of generated images.

## 3.2 Fish Image Generation with Stable Diffusion 2

Through stable diffusion 2 (SD2), we obtain high-quality synthetic marine biology dataset images and their associated latent representations, as shown in Figure 5. Each generated image contains a background, species morphology, and detailed features, thus providing a high-quality dataset for subsequent research. By contrast, images generated by latent consistency model (LCM) [22] often present a single, repetitive background pattern, lacking the complexity and diversity expected of real underwater scenes. They also appear as groups of fish rather than individual fish. The spatial arrangement of fish does not conform to natural behavioral patterns, the occlusion relationships between individuals are handled unnaturally, the symbiotic relationships between different species are difficult to accurately present, and there is a gap in color performance compared to SD2.

In summary, fish recognition tasks often face the challenge of insufficient sample sizes, especially for rare or hard-to-obtain species. As an advanced generative model, Stable Diffusion 2 can produce large numbers of high-quality, diverse synthetic images based on a limited number of samples. These synthetic images can be used to expand training datasets, thereby improving the model generalization and recognition accuracy. Moreover, SD2 enables the generation of training-ready synthetic images without requiring large amounts of annotated data, significantly reducing reliance on manual labeling and saving both time and labor costs. Furthermore, by analyzing the generated synthetic images, researchers can intuitively understand the model's learning process and recognition mechanisms, thus enhancing the model interpretability. This is of great significance for model optimization in both scientific research and practical applications.

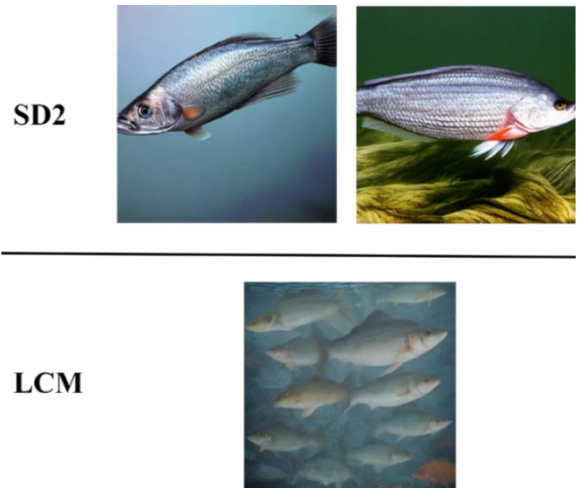

Fig. 5: Example of a Fish Image (Hypophthalmichthys molitrix) Generated by Stable Diffusion 2 and LCM. It is clear that LCM can't provide the single fish and high-quality image either.

## 3.3 Superiority of FishAI 2.0 with the Stable Diffusion 2–CLIP Framework over Other baseline Models

We trained the combined Stable Diffusion 2 and CLIP model, and through iterative parameter tuning, obtained the highest accuracy among all tested settings. We found that the model achieved an accuracy of 91.67% at the family level, 87.58% at the genus level, 85.42% at the species level.

To further validate the robustness of our model FishAI 2.0, we conducted comparison experiments with other commonly used backbone networks, including ViT, BeiT, both of which are recent popular architectures known for their strong performance. CLIP and ViT have already been discussed in previous sections, where their contributions to few-shot learning were highlighted. For scientific rigor, we applied pretraining and parameter optimization (e.g., batch size, number of epochs) to all comparison models to ensure the fairness in performance evaluation. The final results are summarized in Table 4. (ViT, BeiT[23] and ConvNeXt[24] are direct from FishNet paper).

Table 4. Performance comparison of different backbone networks on the family level

| Model | Family |
|---|---|
| FishAI 2.0 | **91.67%** |
| FishAI 1.0 | 79.74% |
| ResNet-50 | 40.37% |
| ViT | 48.40% |
| BeiT | 54.26% |
| ConvNeXt | 60.61% |

The results demonstrate that FishAI 2.0 with SD2-CLIP achieved the highest accuracy (91.67%), confirming the effectiveness of integrating Stable Diffusion 2 with CLIP for enhancing the classification performance. As shown in Table 4, CLIP alone achieved a comparable accuracy of 91.14%, indicating its strong classification capability even after fine-tuning. FishAI 1.0 with ViT reached an accuracy of 79.74%, showing moderate performance but lagging behind multimodal frameworks due to its reliance on unimodal visual information.

Considering the practical significance of taxonomic resolution, we conducted ablation experiments at the species and genus levels, as these categories hold the greatest importance for scientific research and ecological applications. Accurate recognition at these levels directly supports biodiversity monitoring, ecological conservation, and fisheries management, where distinguishing between closely related species is often crucial. As shown in Table 5, the integration of Stable Diffusion 2 with CLIP (SD2-CLIP) leads to consistent improvements at the family and genus levels, with top-1 accuracies of 91.67% and 87.58%, respectively, outperforming the baseline CLIP model.

Table 5. Ablation Experiment Results at different taxonomic levels (acc1 and acc5 means top-1 accuracy and top-5 accuracy, respectively)

| Model | Family(acc1/acc5) | Genus(acc1/acc5) | Species(acc1/acc5) |
|---|---|---|---|
| FishAI 2.0 (SD2 - CLIP) | **91.67%/97.97%** | **87.58%**/95.47% | 85.42%/94.71% |
| CLIP | 91.14%/**97.97%** | 86.36%/**95.76%** | **86.12%/95.23%** |

However, at the species level, FishAI 2.0 with SD2-CLIP achieves a slightly lower top-1 accuracy (85.42%) compared to CLIP (86.12%), while top-5 performance also decreases marginally (94.71% vs. 95.23%). This counterintuitive result may be explained by the trade-off between the minority-class and majority-class performance. As shown in Figure 6, we conduct the experiments at the species level to prove the effectiveness of SD2 for minority classes. When we focus on few-shot learning, FishAI 2.0 with SD2-CLIP outperforms CLIP for classes with the number of training samples less than 100. By dividing the species to large-size species (the number of training images more than 100), medium size species (the training size between 10 and 100), small size species (the training size less than 10), we calculate the average accuracy of the large size species, medium size species, small size species. For the average accuracy of small size species, FishAI 2.0 outperforms CLIP with a large margin, proving our SD2 module is the key part of FishAI 2.0 for few-shot image recognition, where SD2 generates images of minority classes for enlarging the training images. All in all, SD2 in FishAI 2.0 enhances the representation diversity through synthetic augmentation, which benefits few-shot fish image recognition, where only limited samples exist—thereby improving the classification robustness for rare taxa. However, for species with abundant training examples, the synthetic variability may introduce distributional noise, slightly reducing the global discriminability at fine-grained levels. In other words, SD2-CLIP is more effective at handling rare and underrepresented species, which is the focus of this study.

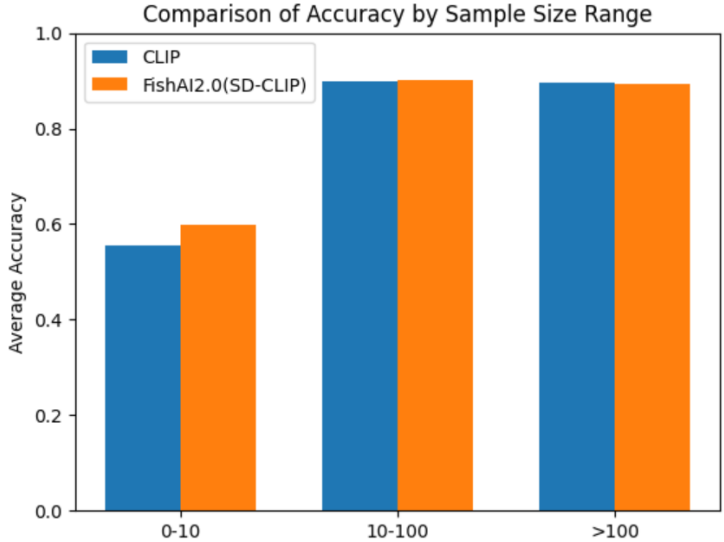

Fig. 6: Comparison of the accuracy for different species of large size (training size more than 100), medium size (training size between 10 and 100), small size (training size less than 10) at the species level. Clearly, it shows that though in large domain CLIP might outperform FishAI 2.0 with SD-CLIP, in medium and small domain, FishAI 2.0 with SD-CLIP outperforms CLIP, demonstrating the effectiveness of FishAI 2.0 for minority classes.

## 3.4 Few-Shot Fish Image Recognition Analysis and Comparison

To evaluate the performance of FishAI 2.0 under few-shot conditions, we compared it with three models CLIP, ViT, and SD-CLIP using 1 to 10 training samples per class. As shown in Fig. 7, the three models exhibit substantial differences in few-shot scenarios. It can be observed that FishAI 2.0 with SD-CLIP demonstrates a dominant performance, achieving an accuracy of 68.5% under the 1-shot condition, which represents improvements of 16.1% and 24.8% over CLIP (52.4%) and ViT (43.7%), respectively. As the number of samples increases, this advantage remains consistent: for 5-shot classification, FishAI 2.0 with SD-CLIP reaches an accuracy of 82.2%, outperforming CLIP by 10.9%; for 10-shot classification, FishAI 2.0 with SD-CLIP achieves an accuracy of 87.4%, exceeding CLIP by 8.4%. This also confirms the robustness of CLIP in few-shot recognition than other backbone models, as it demonstrates strong adaptation to the classes with limited samples—particularly under the 1-shot setting, where it surpasses ViT by 8.7%. However, its relative improvement diminishes as the sample size increases, reflecting the limitations of relying solely on pre-trained knowledge. In contrast, ViT shows the weakest performance, especially under extremely small sample sizes (1–3 samples), highlighting its sensitivity to the data scarcity. Notably, ViT requires at least 5 samples to match the 1-shot accuracy of CLIP. The underlying reason may lie in SD-CLIP's ability to leverage latent codes that provide semantic guidance beyond pixel-level features; under the 1-shot setting, this module contributes approximately 9.2% accuracy improvement.

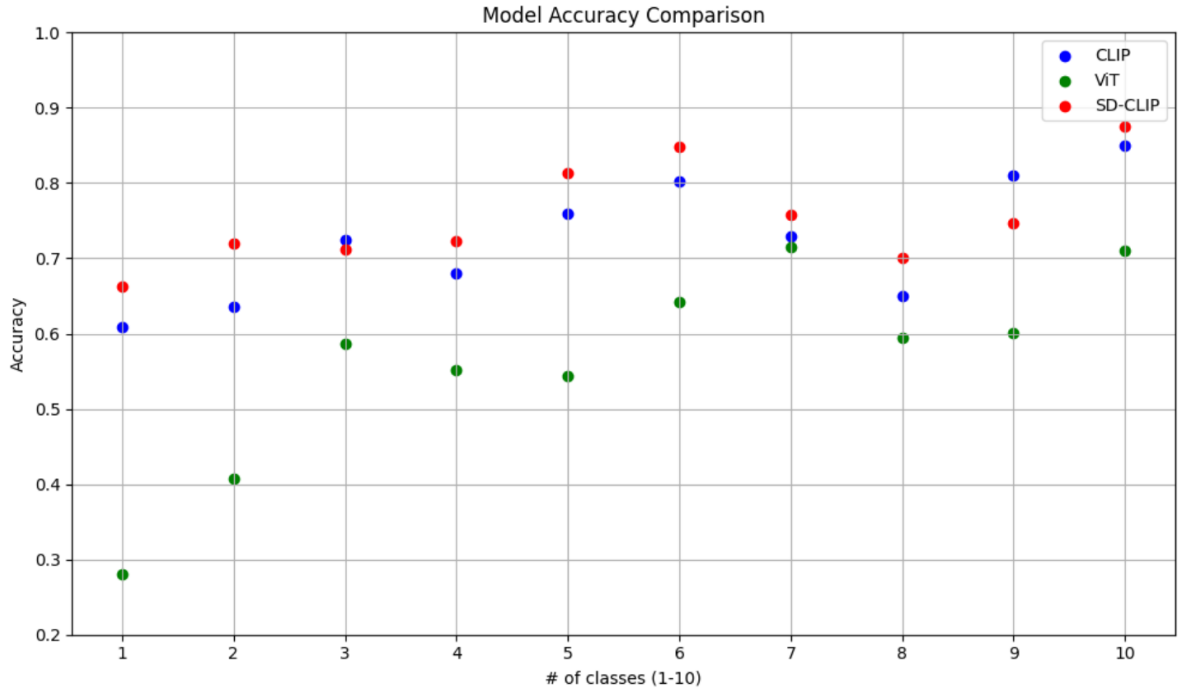

Fig. 7: Accuracy comparison of FishAI 2.0 with SD-CLIP and baseline models under few-shot datasets. Mainly, FishAI 2.0 with SD-CLIP outperformed CLIP and ViT when the amount of training data is small.

More specifically, as shown in Fig. 8 we compare the prediction between FishAI 1.0 and FishAI2.0 for Hypophthalmichthys molitrix, which is rare in FishNet with only a few annotated images. We found that FishAI 2.0 accurately distinguish the invasive fish, while FishAI1.0 cannot output the right result, which show the *Hypophthalmichthys molitrix* as *Coregonus lavaretus*.

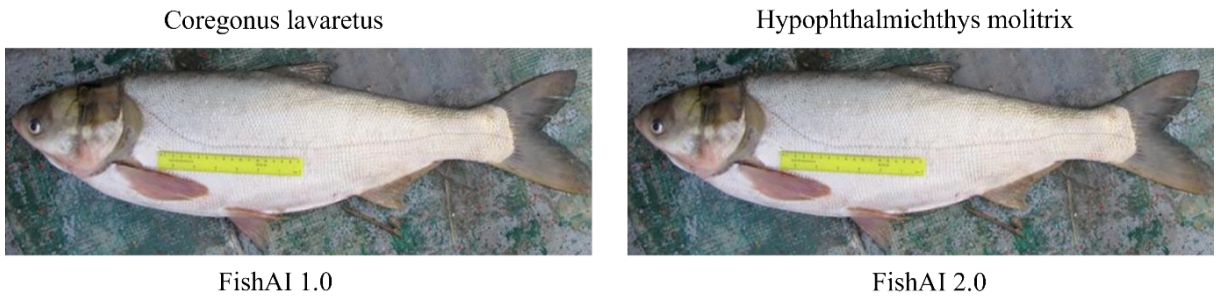

Fig. 8: Prediction result between FishAI 1.0 and FishAI 2.0 for Hypophthalmichthys molitrix, proving that FishAI 2.0 outperforms FishAI 1.0 for identifying the species under few-shot datasets.

Through these experiments, we demonstrate that FishAI 2.0 achieves the progress in few-shot learning. With only 1–3 specimens, it becomes feasible to rapidly archive and identify new species, providing the technical feasibility for monitoring deep-sea and especially rare fish species.

## 4 Conclusion

This study explored the application of multi-modal few-shot learning to marine fish classification by integrating multimodal deep learning with traditional taxonomy. We utilized FishNet, a benchmark dataset with 94,532 annotated images, achieving comprehensive taxonomic coverage from family to species. Methodologically, we designed a novel framework that combines DeepSeek-V3 for high-quality text generation with Stable Diffusion 2 for hierarchical data augmentation, extracting latent encoding to build a multimodal feature space. Coupled with CLIP, this framework significantly improved the classification performance, achieving a Top-1 and Top-5 accuracy of 91.67% and 97.97%, respectively and demonstrating superior robustness under few-shot settings compared to CLIP and ViT alone.

Looking ahead, future work will expand the dataset to broader species coverage, establish international data-sharing standards, and incorporate multimodal sources such as video and sonar. Algorithmic improvements will target extreme few-shot scenarios (1-shot, 0-shot), model compression for mobile deployment, and integration with domain knowledge such as fish ecology and behavior. In terms of applications, customized systems for scientific research, fisheries regulation, aquaculture, and education will be developed, potentially linked with underwater robots, smart fishing nets, and real-time monitoring platforms. In summary, this work represents a foundational step in few-shot learning for marine fish recognition, providing both methodological advances and practical approaches for marine science, ecological conservation, and sustainable development.

## Author contributions

XP and PZ designed this project. CY implemented the methods and performed the analysis. CY, PZ, DSZ, YYW, HS, XP wrote the original manuscript. All authors approved this manuscript.

## Acknowledgements

This work was sponsored by the National Key Research and Development Program of China (2022YFC2804001 and 2024YFC2814401) and the Oceanic Interdisciplinary Program of Shanghai Jiao Tong University (No. SL2022ZD108).

## Conflict of interest

The authors declare there is no conflict of interest.